\documentclass[10pt,twocolumn,letterpaper]{article}

\usepackage[pagenumbers]{wacv} 

\usepackage{graphicx}
\usepackage{wrapfig}
\usepackage{pifont}
\usepackage{cuted} 
\usepackage[table,xcdraw]{xcolor} 

\definecolor{wacvblue}{rgb}{0.21,0.49,0.74}
\usepackage[pagebackref,breaklinks,colorlinks,allcolors=wacvblue]{hyperref}

\def\wacvPaperID{2438} 
\def\confName{WACV}
\def\confYear{2026}

\title{milliMamba: Specular-Aware Human Pose Estimation via Dual mmWave Radar with Multi-Frame Mamba Fusion}

\author{Niraj Prakash Kini$^{\dagger}$, Shiau-Rung Tsai$^{\dagger}$, Guan-Hsun Lin$^{\dagger}$, \\
Wen-Hsiao Peng$^{\dagger}$, Ching-Wen Ma$^{\dagger}$, Jenq-Neng Hwang$^{\ddagger}$\\
${}^{\dagger}$National Yang Ming Chiao Tung University, Taiwan, ${}^{\ddagger}$University of Washington, USA\\
{\tt\small \{nirajnycu.ee06,  mick20001108.cs12, 
abc900203abc.cs12\}@nycu.edu.tw}, \\
{\tt\small wpeng@cs.nycu.edu.tw, machingwen@nycu.edu.tw, hwang@uw.edu}
}

\begin{document}
\maketitle

\begin{strip}
\centering
\includegraphics[width=\textwidth]{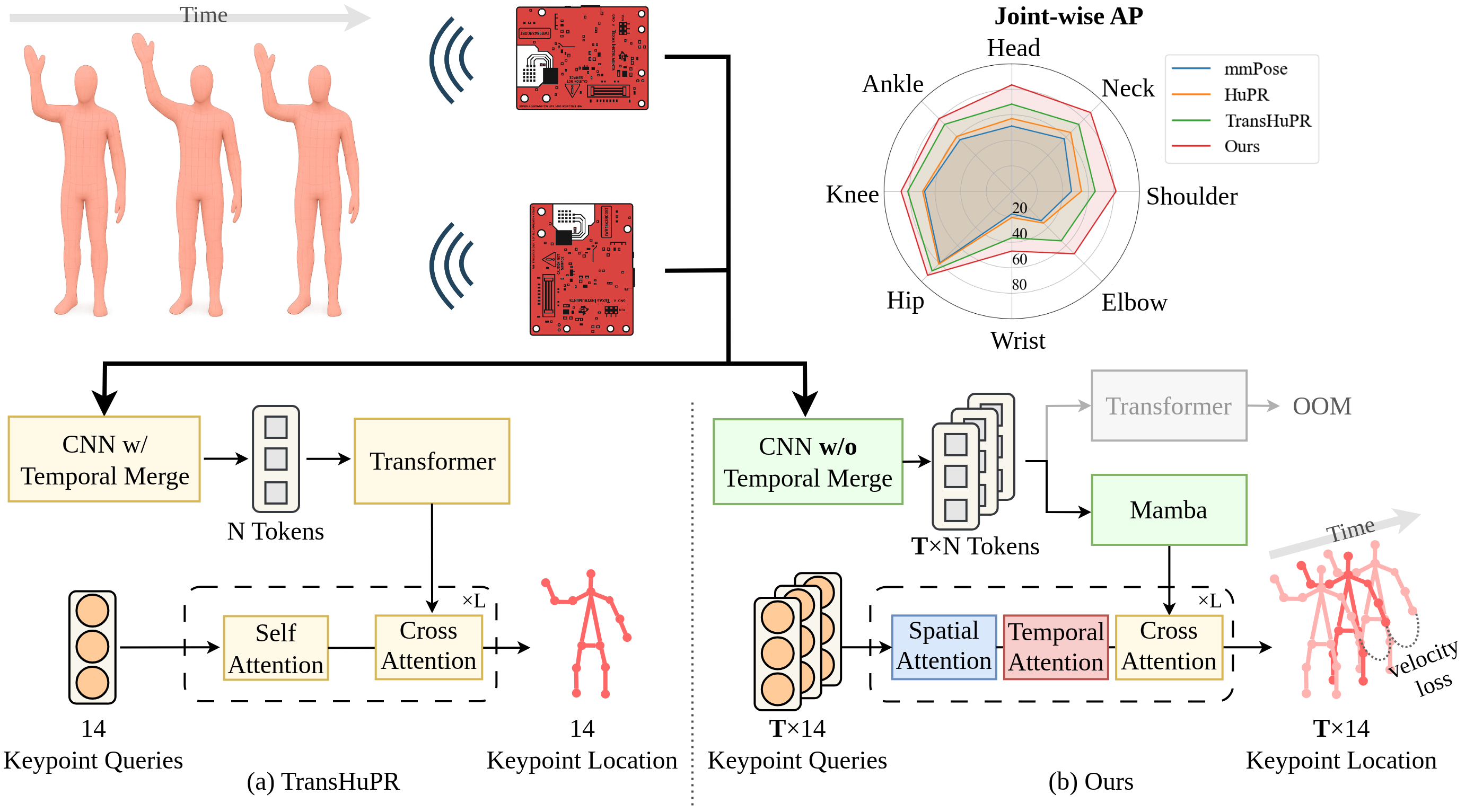}

\captionof{figure}{Our milliMamba performs spatio-temporal modeling across both the feature extraction and decoding stages, addressing a key limitation of TransHuPR~\cite{transhupr}, which models these dependencies only partially. This is made possible by milliMamba’s ability to process a larger number of tokens with a comparable memory footprint, enabling richer temporal context and more accurate pose estimation.}

\label{fig:teaser}
\end{strip}

\begin{table*}[t]
\caption{Comparative analysis of mmWave radar-based HPE approaches.}
\label{tab:comparison_table}
\renewcommand{\arraystretch}{1.1}
\setlength{\tabcolsep}{3pt}
\begin{tabular}{ll|cccc}
\hline
\textbf{Model} &  & \textbf{Input Representation} & \textbf{Prediction Strategy} & \textbf{Early Fuse Temporal} & \textbf{Multi-radar} \\ \hline
RFMamba~\cite{rfmamba} & ICLR'25 & Raw Signal & Many-to-one & \ding{55} & \ding{55} \\
TransHuPR~\cite{transhupr} & BMVC'24 & Point Cloud & Many-to-one & \ding{51} & \ding{51} \\
HuPR~\cite{hupr} & WACV'23 & 4D Heatmap & Many-to-one & \ding{51} & \ding{51} \\
CPFormer~\cite{cpformer} & IEEE Sens.J.'25 & 3D Heatmap & Many-to-one & \ding{51} & \ding{51} \\
RF-Pose~\cite{rfpose} & CVPR'18 & 2D Heatmap & Many-to-many & \ding{55} & WiFi \\ \hline
\textbf{Ours (milliMamba)} &  & 3D Heatmap & Many-to-many & \ding{55} & \ding{51} \\ \hline
\end{tabular}
\end{table*}

\begin{abstract}
Millimeter-wave radar offers a privacy-preserving and lighting-invariant alternative to RGB sensors for Human Pose Estimation (HPE) task. However, the radar signals are often sparse due to specular reflection, making the extraction of robust features from radar signals highly challenging. To address this, we present milliMamba, a radar-based 2D human pose estimation framework that jointly models spatio-temporal dependencies across both the feature extraction and decoding stages. Specifically, given the high dimensionality of radar inputs, we adopt a Cross-View Fusion Mamba encoder to efficiently extract spatio-temporal features from longer sequences with linear complexity. A Spatio-Temporal-Cross Attention decoder then predicts joint coordinates across multiple frames. Together, this spatio-temporal modeling pipeline enables the model to leverage contextual cues from neighboring frames and joints to infer missing joints caused by specular reflections. To reinforce motion smoothness, we incorporate a velocity loss alongside the standard keypoint loss during training. Experiments on the TransHuPR and HuPR datasets demonstrate that our method achieves significant performance improvements, exceeding the baselines by 11.0 AP and 14.6 AP, respectively, while maintaining reasonable complexity. Code: \url{https://github.com/NYCU-MAPL/milliMamba}
\end{abstract}    
\begin{figure*}[htbp]
    \centering
    \includegraphics[width=1.0\textwidth]{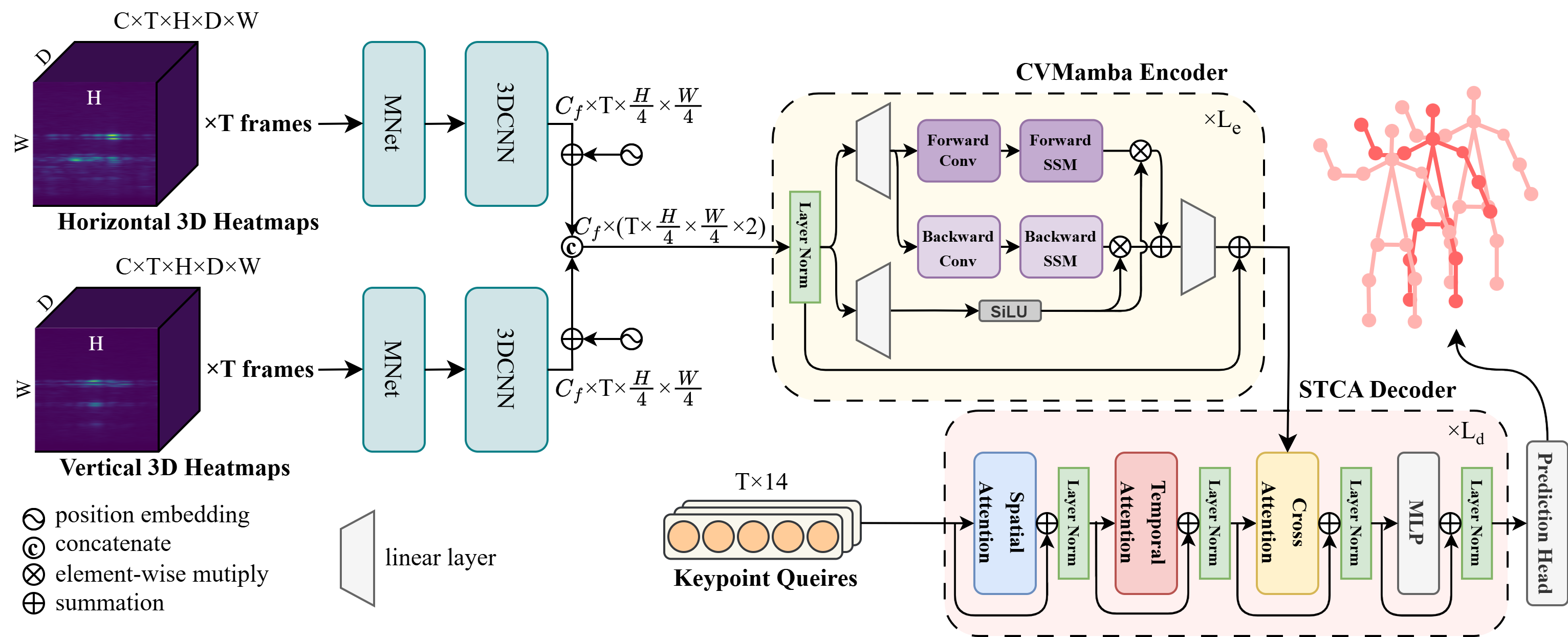}
    \caption{Overview of our milliMamba. The CVMamba encoder first extracts features from dual-view radar inputs. These features are then passed to the Multi-Pose STCA decoder, which progressively refines a set of keypoint queries to produce pose predictions.}
    \label{fig:pipeline}
\end{figure*}

\section{Introduction}
\label{sec:intro}

Millimeter-wave (mmWave) radar-based HPE~\cite{captureHPE, Radhar, mmposeNLP, mmMesh, rtpose, IRUWBRadar, mvdoppler, probradarm3f, rahman2024mmvr, wu2024mmhpe, mueller2025radproposer, huang2025one, engel2025advanced, yan2024indoor, xie2023rpm} has emerged as a compelling alternative to RGB-based pose estimation methods, offering a unique balance of privacy preservation, environmental robustness, and deployment practicality. Despite these advantages, mmWave radar-based HPE remains technically challenging. Due to the specular nature of radar sensing, only body surfaces that reflect signals directly back to the receiver are captured, while others especially small or obliquely oriented joints are often missing. This leads to incomplete observations, making full-body pose reconstruction from single-frame inputs difficult. Additionally, weak reflections from extremities, fluctuations that disrupt temporal consistency, and high sensitivity to subject orientation and sensor placement further hinder estimation accuracy.

To address this issue, we introduce \textit{milliMamba}, an mmWave radar-based 2D HPE framework that incorporates spatial and temporal modeling into both the encoding and decoding stages. This spatio-temporal modeling pipeline enables the model to leverage contextual cues from neighboring frames to infer missing joints caused by specular reflections. Beyond the modeling pipeline itself, we also revisit the signal preprocessing stage to make spatio-temporal modeling more tractable. Instead of constructing computationally expensive 4D heatmaps~\cite{radar_module}, we apply a commonly used 3D Fast Fourier Transform (FFT) to convert raw radar signals into 3D heatmaps. This implementation not only reduces preprocessing overhead but also mitigates the explosion of token counts, making the radar heatmap data easier to handle for downstream modeling.

To this end, our framework integrates two key components: a Mamba-based encoder and an attention-based decoder. The encoder is designed to efficiently process the large token volumes inherent in longer radar sequences, a challenge for prior Transformer-based approaches~\cite{MPTFormer, MetaFi++, IRUWBRadar,tokenpose, lv2025egohand, wang2024gtpt, yataka2024retr, jin2024rodar, jiao2025optimizing, liu2025tcpformer} due to their quadratic complexity. Some methods~\cite{cpformer, transhupr, hupr} attempt to mitigate this by collapsing the temporal dimension early, but such early fusion can compromise the model’s ability to recover missing joints caused by specular reflections.

We adopt a Cross-View Fusion Mamba encoder (CVMamba) that models longer-range spatio-temporal dependencies with linear complexity and effectively fuses dual-radar inputs across frames. To our knowledge, this is the first adaptation of Mamba for cross-view fusion in the radar domain, achieved by modifying the sequential scanning strategy from Vision Mamba~\cite{vim, videomamba}. This design leverages Mamba’s ability to capture dependencies over longer sequences efficiently, making it well-suited for the larger temporal context and multi-view correlations present in dual-radar setups.

The encoded features are passed to a Spatio-Temporal-Cross Attention (STCA) decoder, adapted from a DETR-style architecture~\cite{DETR} to support multi-frame pose prediction. While most prior radar-based HPE methods adopt a multi-frame to single-frame decoding scheme, STCA predicts poses for multiple frames simultaneously, as illustrated in Figure~\ref{fig:teaser}. STCA integrates both spatial attention and temporal attention, enabling it to model spatial relationships within each frame while capturing temporal dependencies across frames. This design offers two benefits: (1) richer supervision across time steps improves pose accuracy, and (2) the model better infers missing joints by leveraging contextual cues from neighboring frames and joints.

We evaluate our method on two mmWave radar-based 2D HPE datasets, TransHuPR~\cite{transhupr} and HuPR~\cite{hupr}. Our method achieves significant performance improvements over the baselines, exceeding them by 11.0 AP and 14.6 AP, respectively, while maintaining an acceptable trade-off between accuracy and complexity.

This work makes the following contributions:
\begin{itemize}
\item We adopt the CVMamba encoder to model longer-range spatio-temporal dependencies with linear complexity while performing cross-view fusion of dual-radar inputs.

\item We propose the STCA decoder, a spatio-temporal cross-attention module that leverages multiple output frames to incorporate additional regularization enhancing pose accuracy and mitigating the effects of missing joints from specular reflections.

\item Extensive experiments show that our proposed milliMamba serves as the new benchmark for the radar-based 2D HPE task on HuPR~\cite{hupr} and TransHuPR~\cite{transhupr} datasets, with a significant leap in results over prior work. 
\end{itemize}

\begin{figure*}[t]
    \centering
    \includegraphics[width=\linewidth]{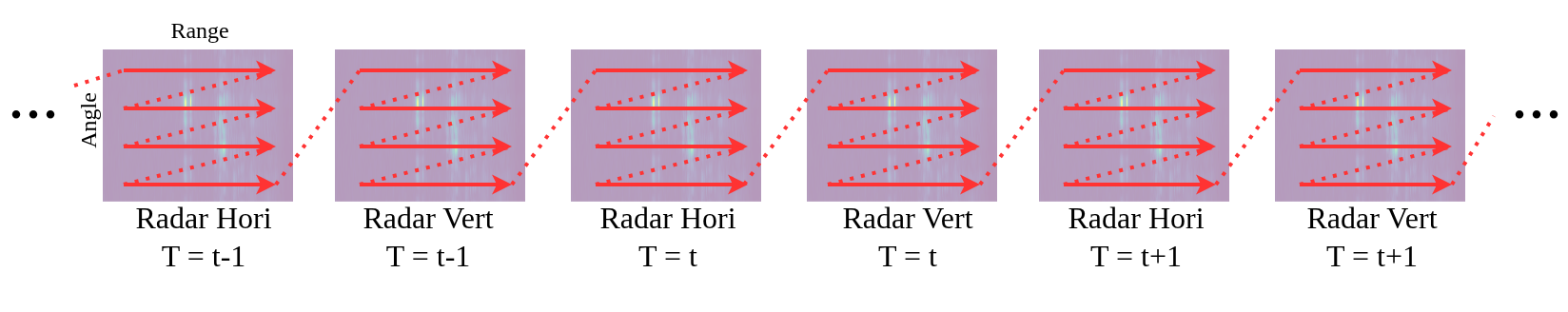}
    \vspace{-25pt}
    \caption{
    Our mamba scanning pattern. Only the forward direction is shown for clarity.
    }
    \label{fig:mamba_scan}
\end{figure*}

\section{Related Work}
\label{sec:relatedwork}


\subsection{CNN- and Transformer-Based mmWave HPE}
Deep learning approaches for mmWave-based human pose estimation (HPE) primarily leverage CNNs and Transformers. CNN-based methods~\cite{mmpose, hupr, mmposeNLP} typically employ 2D or 3D convolutional blocks for feature extraction, followed by fully connected layers for pose prediction. These architectures are effective at capturing multiscale spatial and short-term temporal features, resulting in smooth pose estimations. However, they are often limited in their ability to fuse information from multiple radar sensors.

In contrast, Transformer-based methods~\cite{transhupr, RadarFormer, IRUWBRadar} have demonstrated strong performance in fusing features from multiple radar sources. Their ability to model global dependencies across spatial and temporal dimensions leads to improved pose accuracy. Nevertheless, Transformers generally incur high computational costs, particularly in terms of memory usage and training time.

\subsection{mmWave radar-based HPE}
Table~\ref{tab:comparison_table} compares representative mmWave radar-based HPE approaches across input representation, prediction strategy, temporal fusion, and multi-radar support. RF-Pose~\cite{rfpose} is a WiFi-based HPE that relies on 2D heatmaps, while CPFormer~\cite{cpformer} and HuPR~\cite{hupr} use higher-dimensional heatmaps, and TransHuPR~\cite{transhupr} employs point cloud projections. RFMamba~\cite{rfmamba} using high resolution SFCW radar, uniquely processes raw signals with a many-to-one strategy, though without multi-radar support. Besides RF-Pose~\cite{rfpose}, most prior work adopts many-to-one prediction and relies on early temporal fusion. In contrast, our proposed milliMamba combines a many-to-many prediction strategy with 3D heatmap inputs and multi-radar support, striking a balance between spatial richness and efficiency. Its design inherently captures temporal dependencies without explicit early fusion, allowing superior robustness, efficiency, and accuracy over existing methods.

\subsection{Mamba-Based mmWave HPE: RFMamba}

While Transformers dominate the field, a recent effort has explored state space models (SSMs) for mmWave-based HPE. RFMamba~\cite{rfmamba} is, to date, the only method that applies SSMs to radar-based 3D HPE. Its architecture retains complex-valued representations throughout the network. While RFMamba presents a novel application of SSMs in this domain, its source code is not publicly available, which precludes direct comparison in our experiments.

In contrast, our method directly adopts a vision-domain Mamba architecture~\cite{vim, mamba} to process radar signals in the frequency domain. We treat the real and imaginary components of the complex-valued input as separate feature channels, allowing the model to operate entirely in the real domain. This design avoids complex-valued operations, simplifies implementation, and aligns more naturally with standard vision model architectures.

\section{Proposed Method}
\label{sec:method}

Figures~\ref{fig:pipeline} illustrate the pipeline of our proposed framework, \textit{milliMamba}, which transforms dual-view mmWave radar signals into temporally coherent 2D human poses. We adopt a sliding window strategy centered at the current frame, such that the input includes both past and future radar frames. The architecture consists of three sequential stages: \textbf{(i)} radar pre-processing via a simple 3D Fast Fourier Transform, \textbf{(ii)} a Cross-View Fusion Mamba encoder that fuses temporal dynamics and complementary spatial cues from horizontal and vertical views, and \textbf{(iii)} a Spatio-Temporal-Cross Attention decoder that predicts multi-frame pose sequences within the sliding window. Each component is detailed in the following sections.

\subsection{Radar Pre‑processing}
\label{method:radar_pre}

\paragraph{Input Data Format.}
An FMCW radar produces complex‐valued cubes \(\mathbf{X}\in\mathbb{C}^{12\times128\times256}\), whose three dimensions correspond to 12 virtual-antenna pairs, 128 chirps, and 256 ADC samples. Two radars are mounted orthogonally to capture horizontal and vertical views, and we acquire \(T\) consecutive frames from both sensors in each sliding window~\cite{hupr, transhupr}.

\paragraph{Clutter Removal and Chirp Sub‑sampling.}
First, static clutter is removed by subtracting the mean across chirps. Then, the chirp dimension is uniformly subsampled to 8 chirps per frame to reduce computation while preserving Doppler resolution.

\paragraph{3D Fast Fourier Transform (FFT).}
Finally, we convert each radar cube into a 3D angle-doppler-range heatmap. A 1D FFT is first applied along the ADC‑sample dimension, followed by another along the chirp dimension. To enhance angular resolution, the virtual‑antenna dimension, which originally encodes azimuth and elevation, is zero‑padded from 12 to 64 and then transformed by a third 1D FFT. For a 1D FFT, the transform is
\begin{equation}
\mathbf{Y}(m)=\sum_{n=0}^{N-1}\mathbf{X}(n)\, \exp\!\bigl(-j\,\tfrac{2\pi}{N}\,nm\bigr),
\end{equation}
where \(N\) is the input length and \(n,m\) are the indices of input and output. The resulting tensor \(\mathbf{Y}\in\mathbb{C}^{H\times D\times W} = \mathbb{C}^{64\times 8\times 256}\) organizes the data along \textbf{angle} (\(H\)), \textbf{doppler} (\(D\)), and \textbf{range} (\(W\)).

\begin{figure*}[htbp]
    \centering
    \includegraphics[width=\textwidth]{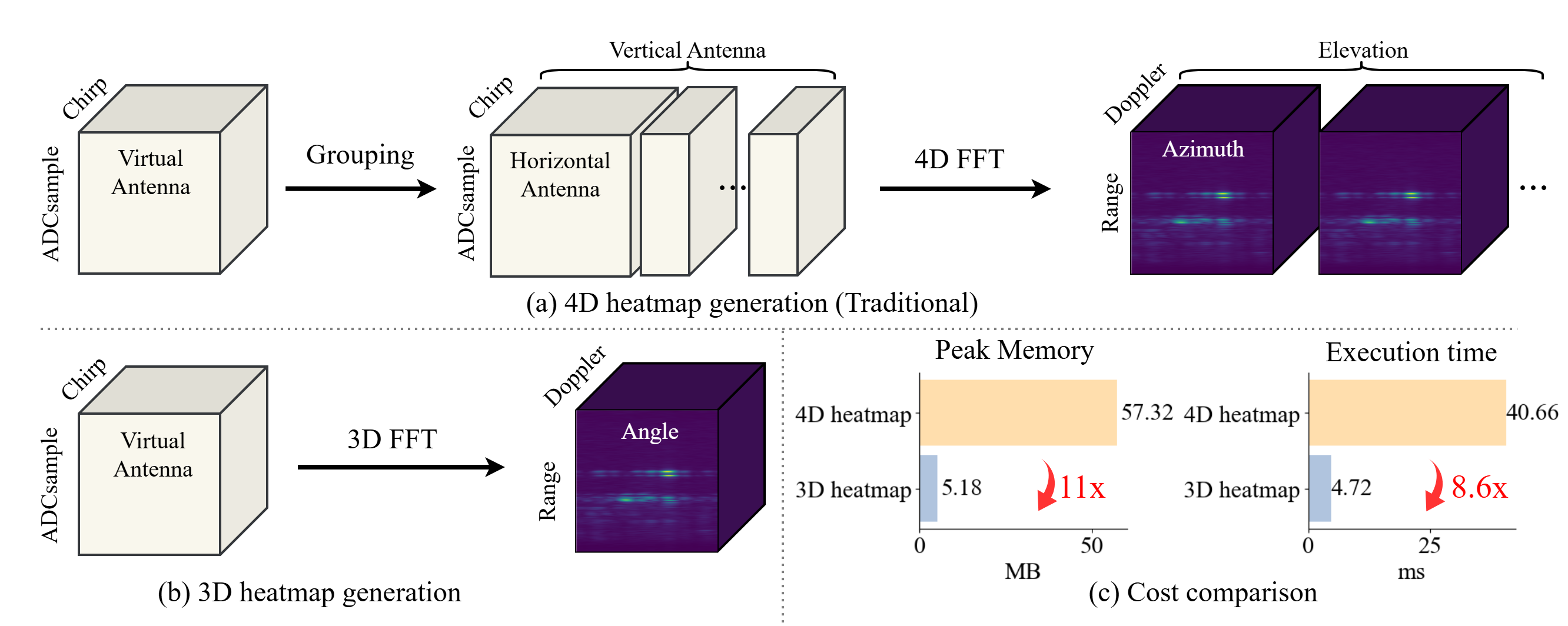}
    \caption{
    Comparison of heatmap generation. (a) The traditional 4D approach~\cite{radar_module} applies separate FFTs for range, doppler, azimuth, and elevation after antenna grouping. (b) Our 3D pipeline performs a unified spatial FFT without grouping, yielding a compact representation. (c) Cost comparison between 4D and 3D heatmaps, showing 11× reduction in memory and 8.6× reduction in latency.
    }
    \label{fig:heatmap_compare}
\end{figure*}

Figure~\ref{fig:heatmap_compare} illustrates the summary of the radar preprocessing pipeline. The 3D heatmap generation simplifies the process and is far more efficient, cutting memory usage by 11× and latency by 8.6× compared to the conventional 4D approach. 

\subsection{Cross‑View Fusion Mamba Encoder (CVMamba)}
\label{method:CVMamba}


The preprocessed radar data from Sec.~\ref{method:radar_pre} are stacked across \(T\) frames and split into real and imaginary components to form a two‑channel tensor of shape \(C\times T\times H\times D\times W\) with \(C=2\). Following HuPR~\cite{hupr}, we process the horizontal (\(h\)) and vertical (\(v\)) views in two parallel branches. Each branch starts with an MNet~\cite{mnet} block that merges the doppler dimension, then passes through three residual 3D convolutions and two down‑sampling layers, reducing the angle \(H\) and range \(W\) resolutions by \(4\times\). This produces feature maps \(\mathbf{F}_h,\mathbf{F}_v \in\mathbb{R}^{C_f\times T\times \tfrac{H}{4}\times \tfrac{W}{4}}\).

\paragraph{Cross‑View Fusion.}
We first add separate learnable positional embeddings \(\mathbf{P}_h\) and \(\mathbf{P}_v\) to the horizontal and vertical feature maps, encoding angle, and range dimension. The two views are then concatenated to form the encoder input \(\mathbf{F}=[\mathbf{F}_h;\mathbf{F}_v]\in\mathbb{R}^{C_f\times T\times \tfrac{H}{4}\times \tfrac{W}{4}\times2}\). 

\paragraph{Scanning Order.}
We convert \(\mathbf{F}\) into a 1D sequence by a zigzag scan across \textit{range} \(\rightarrow\) \textit{angle} \(\rightarrow\) \textit{view} (\(h\rightarrow v\)) \(\rightarrow\) \textit{frame} shown in  Figure~\ref{fig:mamba_scan}. This sequence is processed by two independent SSM branches in the forward and backward directions to provide bidirectional context.

\paragraph{Vision Mamba Encoder.~\cite{vim}}
The token sequence is processed by a stack of Vision Mamba layers. Each layer integrates a gating mechanism, two directional SSMs, and residual connections to efficiently capture long-range dependencies. Each SSM updates the hidden state \(h_{t+1}\) as:
\begin{equation}
\mathbf{h}_{t+1}=A\,\mathbf{h}_t + B\,u_t,\qquad
\mathbf{y_t} = C\,\mathbf{h}_t + D\,u_t,
\end{equation}
where \(u_t,y_t\) denote the input token and the output token at time step \(t\), and \(A,B,C,D\) are layer-specific learnable parameters. Stacking \(L_e\) such Vision Mamba layers enables the model to extract global context across long sequences while preserving linear complexity. Through this design, our encoder captures features across multiple frames and multiple views, enabling efficient global context modeling.

\subsection{Multi‑Pose Spatio‑Temporal-Cross Attention Decoder (STCA)}
\label{method:MPST}

To predict human poses from the encoded radar representations, we propose a Multi-Pose Spatio-Temporal Cross-Attention (STCA) decoder that leverages learnable keypoint queries to jointly model spatial structure, temporal dynamics, and encoder–decoder interactions.

The decoder operates on a fixed set of \(J\times T\) learnable keypoint queries \(\{\mathbf{q}_{f,j}\}\) where \(f\in\{0,\dots,T-1\}\) indexes frames and \(j\in\{0,\dots,J-1\}\) enumerates joints. Each decoder layer includes \textit{Spatio‑Temporal Attention}, \textit{Cross‑Attention}, and a position‑wise \textit{MLP}. This layer is stacked \(L_d\) times, allowing queries to iteratively refine their representations.

\paragraph{Keypoint Query Embedding.}
Inspired by the object query strategy of DETR~\cite{DETR}, each keypoint query represents a particular joint in a specific frame. All \(J \times T\) keypoint queries are learnable and serve as the decoding tokens for pose prediction.

\paragraph{Spatio‑Temporal Attention Module.}
Each decoder layer begins with two self‑attention blocks that operate directly on the keypoint queries. Spatial Attention is applied first, operating within each frame to aggregate the \(J\) joints and capture inter-joint relationships. Temporal Attention attends to the same joint across all \(T\) frames, thereby enforcing motion consistency.
\begin{align}
\mathbf{q}'_{f,\cdot} &= \text{SA}\bigl(\mathbf{q}_{f,\cdot}\bigr)
   = \operatorname{softmax}\!\bigl(Q_fK_f^{\!\top}/\sqrt{d}\bigr)V_f, \\
\mathbf{q}''_{\cdot,j} &= \text{TA}\bigl(\mathbf{q}'_{\cdot,j}\bigr)
   = \operatorname{softmax}\!\bigl(Q_jK_j^{\!\top}/\sqrt{d}\bigr)V_j.
\end{align}

\paragraph{Cross‑Attention to Encoder Features.}
The updated keypoint queries attend to encoder features \(\mathbf{{F}^{'}}\) through standard cross‑attention. This mechanism enables the model to utilize contextual information from all frames, improving its ability to estimate missing keypoints.
\begin{equation}
\widetilde{\mathbf{q}}_{f,j}  = \operatorname{CrossAttn}\!\bigl(\mathbf{{q}^{''}}_{f,j}, \mathbf{{F}^{'}}\bigr).
\end{equation}

Finally, each refined query \(\widetilde{\mathbf{q}}_{f,j}\) is passed through a prediction head to produce 2D keypoint coordinates, generating a sequence of \(T\) pose estimates.

By iteratively refining keypoint queries through spatio-temporal self-attention and cross-attention with encoder features, the STCA decoder produces temporally consistent 2D pose sequences, effectively capturing both per-frame joint relations and motion continuity.

\begin{table*}
\caption{Comparison of model performance and complexity across methods on the TransHuPR dataset~\cite{transhupr}. The complexity excludes radar signal preprocessing.}
\label{tab:main_comparison_transhupr}
\centering
\resizebox{\linewidth}{!}{
\begin{tabular}{l|ccc|cccccccc|ccc}
\hline
\textbf{Method} & \multicolumn{3}{c|}{\textbf{Complexity}} & \multicolumn{8}{c|}{\textbf{Joint-wise AP}} & \multicolumn{3}{c}{\textbf{Overall AP}} \\
 & MACs & Params & Mem & Head & Neck & Shoulder & Elbow & Wrist & Hip & Knee & Ankle & AP & AP$^{50}$ & AP$^{75}$ \\ \hline
mmPose~\cite{mmpose} & \textbf{85.5 M} & 15.0 M & \textbf{67.2 MB} & 51.2 & 58.2 & 46.8 & 32.6 & 17.7 & 79.5 & 68.3 & 57.3 & 48.4 & 88.4 & 47.4 \\
HuPR~\cite{hupr} & 68.6 G & 35.5 M & 339.7 MB & 57.1 & 65.3 & 54.6 & 35.2 & 20.6 & 80.8 & 69.8 & 60.9 & 51.5 & 89.5 & 53.7 \\
TransHuPR~\cite{transhupr} & 5.8 G & 5.3 M & 230.8 MB & 68.4 & 74.3 & 65.4 & 54.9 & 36.5 & 88.3 & 81.5 & 74.3 & 67.5 & 96.9 & 76.7 \\
\rowcolor[HTML]{E2FEE2} 
Ours & 34.4 G & \textbf{4.0 M} & 224.1 MB & \textbf{83.5} & \textbf{87.4} & \textbf{81.7} & \textbf{69.3} & \textbf{46.9} & \textbf{93.2} & \textbf{86.7} & \textbf{80.6} & \textbf{78.5} & \textbf{98.7} & \textbf{89.3} \\ \hline
\end{tabular}
}
\end{table*}

\begin{table*}
\caption{Comparison of model performance and complexity across methods on the HuPR dataset~\cite{hupr}. The complexity excludes radar signal preprocessing.}
\label{tab:main_comparison_hupr}
\centering
\resizebox{\linewidth}{!}{
\begin{tabular}{l|ccc|cccccccc|ccc}
\hline
\textbf{Method} & \multicolumn{3}{c|}{\textbf{Complexity}} & \multicolumn{8}{c|}{\textbf{Joint-wise AP}} & \multicolumn{3}{c}{\textbf{Overall AP}} \\
 & MACs & Params & Mem & Head & Neck & Shoulder & Elbow & Wrist & Hip & Knee & Ankle & AP & AP$^{50}$ & AP$^{75}$ \\ \hline
mmPose~\cite{mmpose} & \textbf{85.5 M} & 15.0 M & \textbf{67.2 MB} & 56.1 & 60.9 & 40.6 & 24.9 & 14.2 & 63.2 & 58.6 & 56.1 & 41.4 & 79.4 & 38.3 \\
HuPR~\cite{hupr} & 68.6 G & 35.5 M & 339.7 MB & 77.5 & 81.9 & 70.3 & 45.5 & 22.3 & 88.1 & 82.2 & 73.1 & 63.4 & 97.0 & 74.0 \\
TransHuPR~\cite{transhupr} & 5.8 G & 5.3 M & 230.8 MB & 77.1 & 78.6 & 63.2 & 55.6 & 44.9 & 84.5 & 83.6 & 80.0 & 69.4 & 95.1 & 79.9 \\
\rowcolor[HTML]{E2FEE2} 
Ours & 34.4 G & \textbf{4.0 M} & 224.1 MB & \textbf{90.0} & \textbf{91.8} & \textbf{83.2} & \textbf{75.2} & \textbf{59.5} & \textbf{94.3} & \textbf{93.6} & \textbf{89.3} & \textbf{84.0} & \textbf{98.5} & \textbf{94.9} \\ \hline
\end{tabular}
}
\end{table*}

\begin{table}
\caption{Ablation study on the impact of different input representations.}
\label{tab:ablation_input_modality_AP}
\centering
\begin{tabular}{c|ccc}
\hline
\textbf{Input Representation} & AP & AP$^{50}$ & AP$^{75}$ \\ \hline
density map & 58.5 & 92.5 & 62.7 \\
4D FFT & 72.0 & 97.3 & 81.8 \\
3D FFT & \textbf{74.5} & \textbf{98.5} & \textbf{84.7} \\ \hline
\end{tabular}
\end{table}

\subsection{Training Objectives}
We employ two losses. First, the Object Keypoint Similarity \(L_{\text{oks}}\) penalizes mismatches between predicted and ground-truth locations. Second, the velocity loss \(L_{\text{vel}}\) encourages temporal smoothness by minimizing the error between predicted and ground-truth joint velocities~\cite{smoothnet}. We define the predicted velocity of joint \(j\) at frame \(f\) as the difference between its positions at consecutive frames: \(\widehat{\mathbf{v}}_{f,j}=\widehat{\mathbf{p}}_{f+1,j}-\widehat{\mathbf{p}}_{f,j}\) and \(\mathbf{v}_{f,j}\) is computed from ground-truth keypoint positions. The velocity loss is computed as:
\begin{equation}
L_{\text{vel}}=\frac{1}{(T-1)\,J}\sum_{f=1}^{T-1}\sum_{j=1}^{J} \bigl\lVert\widehat{\mathbf{v}}_{f,j}-\mathbf{v}_{f,j}\bigr\rVert_2^2.
\end{equation}

The overall training objective is \(L = L_{\text{oks}} + \lambda_{\text{vel}}\, L_{\text{vel}},\) where \(\lambda_{\text{vel}}\) balances between pose accuracy and temporal consistency. At inference time, only the central frame prediction within each window is retained, while ensuring that the output reflects the most accurate estimate.

\begin{table}
\caption{Ablation study on the multi-pose output mechanism.}
\label{tab:ablation_one_or_nine_pose}
\centering
\begin{tabular}{c|ccc}
\hline
\textbf{Prediction Strategy} & AP & AP$^{50}$ & AP$^{75}$ \\ \hline
Many-to-one & 70.4 & 97.0 & 81.0 \\
Many-to-many & \textbf{74.5} & \textbf{98.5} & \textbf{84.7} \\ \hline
\end{tabular}
\end{table}
\begin{table*}
\caption{Impact of input sequence length ($T$) on pose estimation performance. We investigate the effect of varying $T$ to understand how temporal context contributes to accuracy.}
\label{tab:ablation_num_of_frames_and_pose}
\centering
\resizebox{\linewidth}{!}{
\begin{tabular}{c|ccc|cccccccc|ccc}
\hline
 & \multicolumn{3}{c|}{\textbf{Complexity}} & \multicolumn{8}{c|}{\textbf{Joint-wise AP}} & \multicolumn{3}{c}{\textbf{Overall AP}} \\
\textbf{T} & MACs & Params & Mem & Head & Neck & Shoulder & Elbow & Wrist & Hip & Knee & Ankle & AP & AP$^{50}$ & AP$^{75}$ \\ \hline
3 & 5.6 G & 3.2 M & 44.7 MB & 73.4 & 78.6 & 69.5 & 53.9 & 32.6 & 88.0 & 79.6 & 71.9 & 66.9 & 95.7 & 75.0 \\
5 & 9.4 G & 3.2 M & 62.3 MB & 78.3 & 81.9 & 73.3 & 58.3 & 35.8 & 89.5 & 82.1 & 75.0 & 70.3 & 97.1 & 79.5 \\
7 & 13.3 G & 3.2 M & 86.6 MB & 79.6 & 84.0 & 75.4 & 61.5 & 39.3 & 91.0 & 84.0 & 77.7 & 72.9 & 97.3 & 83.0 \\
9 & 17.3 G & 3.2 M & 121.2 MB & 80.6 & 84.6 & 77.7 & 63.7 & 41.0 & 91.5 & 84.4 & 78.6 & 74.5 & 98.5 & 84.7 \\
11 & 21.4 G & 3.2 M & 163.5 MB & 79.6 & 83.9 & 78.2 & 64.9 & 42.0 & 91.9 & 85.1 & 78.8 & 75.2 & 98.4 & 86.2 \\
13 & 25.6 G & 3.2 M & 212.1 MB & 81.2 & 85.0 & 78.8 & 65.6 & 44.2 & 91.8 & 85.5 & 79.1 & 75.8 & 98.5 & 85.9 \\
15 & 29.9 G & 3.2 M & 269.3 MB & 81.7 & 85.9 & 80.4 & 66.8 & 45.1 & 92.6 & 86.4 & 80.8 & 77.1 & 98.7 & 87.9 \\ \hline
\end{tabular}
}
\end{table*}

\section{Experimental Results}
\label{sec:result}


\subsection{Implementation Details}
\label{sec:implementation_details}

Our main method follows the default configuration described below. The model takes input from two radar sensors, each capturing a sequence of $T = 9$ frames. It outputs 9 consecutive pose predictions, from which only the center pose is used during inference. The model is trained using the Adam optimizer with a learning rate of 0.00005, a batch size of 8, and a weight decay of 0.0001. The velocity supervision term in the loss function is weighted by $\lambda_{\text{vel}} = 0.05$. All experiments are conducted on a single NVIDIA Tesla V100 GPU.

\subsection{Datasets and Evaluation Metrics}
\label{sec:datasets_evaluation_metrics}

We evaluate our method on two benchmark mmWave radar-based 2D HPE datasets. 


\textbf{TransHuPR Dataset}~\cite{transhupr} comprises 440 sequences, totaling over 7 hours of video recorded from 22 subjects. It consists exclusively of fast and dynamic actions, presenting a considerable challenge due to the high motion complexity and diversity. We follow the original data split protocol: 352 sequences are used for training, 44 for validation, and 44 for testing.

\textbf{HuPR Dataset}~\cite{hupr} contains 235 sequences, amounting to approximately 4 hours of video from 6 subjects. In contrast to TransHuPR, the majority of actions in this dataset are relatively static. We adopt the same split protocol: 193 sequences for training, 21 for validation, and 21 for testing.

For evaluation, we use Average Precision (AP) based on Object Keypoint Similarity (OKS)~\cite{ap_metric}. The overall AP is computed by averaging OKS over 10 thresholds uniformly spaced from 0.50 to 0.95. We also report AP$^{50}$ and AP$^{75}$, which correspond to OKS thresholds of 0.50 and 0.75, representing loose and strict matching criteria, respectively.

\subsection{Performance Comparison}
\label{sec:performance_comparison}

Table~\ref{tab:main_comparison_transhupr} presents a comparison between our method and existing radar-based HPE approaches on the TransHuPR dataset. Our method consistently outperforms all baselines across all AP metrics, including both overall and joint-wise AP, while also maintaining a relatively compact model size. In particular, compared to the baseline \textit{TransHuPR}~\cite{transhupr}, our approach achieves a substantial improvement of 11.0 AP. Notably, for the most challenging joint-wrist, which is both fast-moving and often affected by specular reflection, our model achieves an AP of 46.9, demonstrating its robustness in inferring joints with high uncertainty. Our model consistently improves accuracy across all keypoints, indicating reliable performance over the entire body structure. Figure~\ref{fig:vis_transhupr} presents qualitative results on the TransHuPR dataset. Additional examples are in the supplementary material.

Table~\ref{tab:main_comparison_hupr} shows a similar trend on the HuPR dataset. Our method achieves up to 84.0 AP, indicating high prediction accuracy on relatively static actions. Compared to the baseline \textit{HuPR}~\cite{hupr} which requires 68.6 GMACs and 35.5 M parameters, our method achieves higher accuracy with only 34.4 GMACs and 4.0 M parameters, highlighting the computational efficiency of our design. All baseline methods adopt a single-frame output strategy, whereas our milliMamba employs a multi-frame output strategy.

While \textit{TransHuPR} model exhibits lower MACs, our method offers significantly higher accuracy with only a moderate increase in compute. This demonstrates a favorable balance between efficiency and performance.

\subsection{Ablation Studies}
\label{sec:ablation}

We conduct a series of ablation studies on the TransHuPR dataset to evaluate the impact of key design choices. Unless otherwise specified, all experiments are performed using a simplified variant of our model that employs only a single vertical radar. In this setting, the second CNN branch is removed, while the rest of the architecture remains unchanged. In addition, the number of default input frames \(T=9\).

\paragraph{Input Representation.} As shown in Table~\ref{tab:ablation_input_modality_AP}, we compare different radar pre-processing strategies. The \textit{density map}, a 2D projection derived from the 3D point cloud representation~\cite{transhupr}, yields the lowest performance, underscoring the limitations of compressing 3D point clouds into 2D representations. Heatmap-based representation performs better, and the \textit{3D FFT}-based heatmap, which omits elevation padding and one FFT dimension used in 4D FFT, achieves comparable accuracy. Moreover, as illustrated in Figure~\ref{fig:heatmap_compare}(c), 3D FFT incurs significantly lower preprocessing cost than 4D FFT, thereby validating our design choice.


\begin{table}
\caption{Effect of radar configuration on pose estimation performance. We compare three setups: horizontal-only (Hori), vertical-only (Vert), and dual-radar (Hori+Vert).}
\label{tab:ablation_num_of_radars}
\centering
\resizebox{\linewidth}{!}{
\begin{tabular}{c|ccc|ccc}
\hline
\textbf{Radar Used} & \multicolumn{3}{c|}{\textbf{Complexity}} & \multicolumn{3}{c}{\textbf{Overall AP}} \\
 & MACs & Params & Mem & AP & AP$^{50}$ & AP$^{75}$ \\ \hline
Hori & 17.3 G & 3.2 M & 121.2 MB & 67.3 & 95.8 & 75.0 \\
Vert & 17.3 G & 3.2 M & 121.2 MB & 74.5 & 98.5 & 84.7 \\
Hori+Vert & 34.4 G & 4.0 M & 224.1 MB & 78.5 & 98.7 & 89.3 \\ \hline
\end{tabular}
}
\end{table}

\begin{table}
\caption{Comparison of Transformer and Mamba encoders with 3-frame radar inputs. Transformer runs out-of-memory on our hardware when trained with longer sequences.}
\label{tab:ablation_transformer_mamba}
\centering
\resizebox{\linewidth}{!}{
\begin{tabular}{c|ccc|ccc}
\hline
\textbf{Encoder} & \multicolumn{3}{c|}{\textbf{Complexity}} & \multicolumn{3}{c}{\textbf{Overall AP}} \\
 & MACs & Params & Mem & AP & AP$^{50}$ & AP$^{75}$ \\ \hline
Transformer & 14.9 G & 3.9 M & 610.3 MB & 65.4 & 95.5 & 73.5 \\
Mamba & 5.6 G & 3.2 M & 44.7 MB & 66.9 & 95.7 & 75.0 \\ \hline
\end{tabular}
}
\end{table}



\begin{figure*}[t]
    \centering
    \includegraphics[width=\textwidth, height=11.2cm]{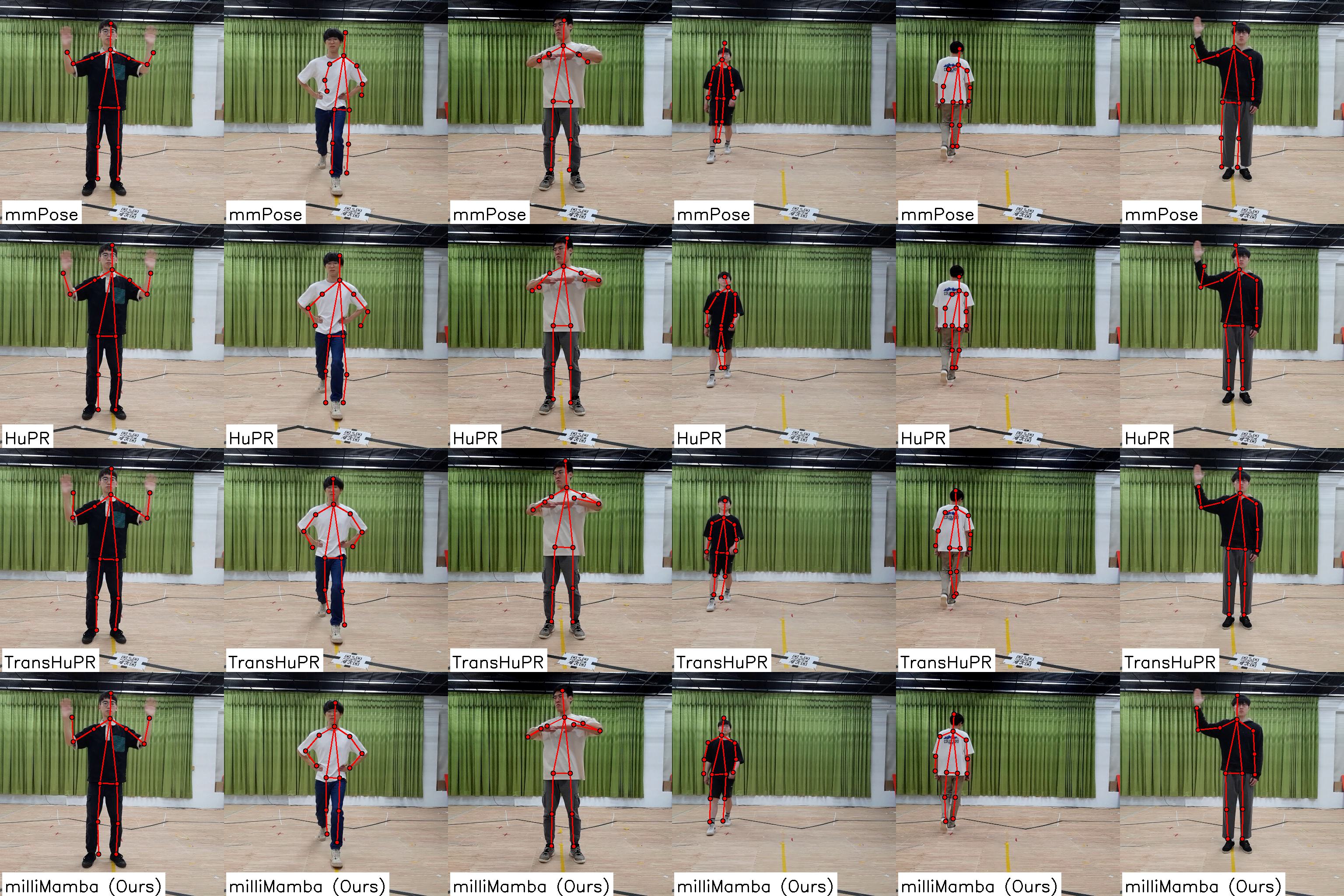}
    \caption{Qualitative results on TransHuPR~\cite{transhupr} dataset.}
    \label{fig:vis_transhupr}
\end{figure*}

\paragraph{Multi-Pose Output Mechanism.} Table~\ref{tab:ablation_one_or_nine_pose} compares our model, \textit{Many-to-many}, with a simplified variant, \textit{Many-to-one}, where the decoder is replaced by a vanilla Transformer~\cite{transformer} that receives only center-pose keypoint queries and predicts a single pose. Our Multi-Pose STCA Decoder achieves a 4.1 AP improvement in overall accuracy. Although our method predicts only the center pose at inference time, the prediction is guided by joint features from different time steps, enabling the model to infer missing or weakly reflected joints using rich spatial-temporal context. More details in supplementary materials.

\paragraph{Effect of Input Sequence Length.} Table~\ref{tab:ablation_num_of_frames_and_pose} presents the impact of varying the number of input frames \(T\) on pose estimation accuracy. The results indicate that increasing \(T\) consistently improves performance, particularly for joints affected by rapid motion or frequently missing due to specular reflection, such as the wrist and elbow. Considering the trade-off between accuracy and computational cost, we adopt \(T=9\) as the default setting.

\paragraph{Number of Radars.} Table~\ref{tab:ablation_num_of_radars} explores different radar configurations. Surprisingly, even using a single vertical radar achieves competitive performance, showcasing the practicality of single-radar setups for real-world deployment. While single-radar systems are simpler and more cost-effective, the dual-radar configuration offers additional gains by compensating for the limited elevation resolution inherent to mmWave radar sensors.

\paragraph{Transformer vs Mamba.} Table~\ref{tab:ablation_transformer_mamba} compares the encoder performance of Transformer and Mamba. Due to the higher memory demands of the Transformer, we were only able to conduct experiments with \(T=3\) frames to avoid out-of-memory issues. The results show that Mamba achieves 1.5 AP higher than the Transformer, indicating comparable accuracy. More importantly, given limited memory resources, Mamba offers a practical solution that scales effectively to longer input sequences. The Transformer requires 14.9 GMACs at \(T=3\), computationally comparable to Mamba at \(T=7\) and \(T=9 \) where our model achieves much higher AP, highlighting Mamba’s superior scalability at similar computational cost.


\section{Conclusion}
\label{sec:conclusion}


We propose \textit{milliMamba}, a dual-radar architecture for human pose estimation that tackles sparse reflections and high-dimensional radar signals. By combining efficient input processing with multi-frame sequence modeling, milliMamba captures rich spatio-temporal context for robust pose prediction. Experiments on two benchmarks show state-of-the-art performance with competitive efficiency. Future work will explore multi-person and cross-environment scenarios while further reducing computational cost.

\section{Acknowledgement}
\label{sec:acknowledgement}
This work is supported by National Science and Technology Council (NSTC), Taiwan, under Grants 113-2634-F-A49-007-, 112-2221-E-A49-092-MY3, and 114-2221-E-A49-035-MY3. We thank to National Center for High-performance Computing (NCHC) for providing computational and storage resources.
{
    \small
    \bibliographystyle{ieeenat_fullname}
    \bibliography{main}

@String(BMVC= {Brit. Mach. Vis. Conf.})

@String(AAAI = {AAAI})

@String(BMVC  =	{BMVC})

@ARTICLE{RadarFormer,
  author={Zheng, Zhijie and Zhang, Diankun and Liang, Xiao and Liu, Xiaojun and Fang, Guangyou},
  journal={IEEE Transactions on Neural Networks and Learning Systems}, 
  title={RadarFormer: End-to-End Human Perception With Through-Wall Radar and Transformers}, 
  year={2023},
  volume={},
  number={},
  pages={1-15},
  month={Sep}
}

@ARTICLE{MPTFormer,
  author={Chen, Lin and Guo, Xuemei and Wang, Guoli},
  journal={IEEE Sensors Journal}, 
  title={MPTFormer: Towards Robust Arm Gesture Pose Tracking Using Dual-view Radar System}, 
  year={2023},
  volume={23},
  month={Nov}}

@ARTICLE{MetaFi++,
  author={Zhou, Yunjiao and Huang, He and Yuan, Shenghai and Zou, Han and Xie, Lihua and Yang, Jianfei},
  journal={IEEE Internet of Things Journal}, 
  title={MetaFi++: WiFi-Enabled Transformer-Based Human Pose Estimation for Metaverse Avatar Simulation}, 
  year={2023},
  volume={10},
  number={16},
  pages={14128-14136},
  month={Mar},
}

@InProceedings{captureHPE,  
author={Li, Guangzheng and Zhang, Ze and Yang, Hanmei and Pan, Jin and Chen, Dayin and Zhang, Jin},  
booktitle={2020 IEEE International Conference on Pervasive Computing and Communications Workshops (PerCom Workshops)},   
title={Capturing Human Pose Using mmWave Radar},  
month = {Mar},
year={2020}, 
}

@inproceedings{Radhar,
author = {Singh, Akash Deep and Sandha, Sandeep Singh and Garcia, Luis and Srivastava, Mani},
title = {RadHAR: Human Activity Recognition from Point Clouds Generated through a Millimeter-Wave Radar},
booktitle = {Proceedings of the 3rd ACM Workshop on Millimeter-Wave Networks and Sensing Systems},
month = {Oct},
year = {2019}
}

@ARTICLE{mmposeNLP,
  author={Sengupta, Arindam and Cao, Siyang},
  journal={IEEE Transactions on Neural Networks and Learning Systems}, 
  title={mmPose-NLP: A Natural Language Processing Approach to Precise Skeletal Pose Estimation Using mmWave Radars}, 
  year={2022},
  month={Mar},
  pages={267–281}
}

@inproceedings{mmMesh,
author = {Xue, Hongfei and Ju, Yan and Miao, Chenglin and Wang, Yijiang and Wang, Shiyang and Zhang, Aidong and Su, Lu},
title = {mmMesh: Towards 3D Real-Time Dynamic Human Mesh Construction Using Millimeter-Wave},
booktitle={Proceedings of the 19th Annual International Conference on Mobile Systems, Applications, and Services (MobiSys)},
month = {June},
year = {2021},
}

@ARTICLE{IRUWBRadar,
  author={Kim, Gon Woo and Lee, Sang Won and Son, Ha Young and Choi, Kae Won},
  journal={IEEE Access}, 
  title={A Study on 3D Human Pose Estimation Using Through-Wall IR-UWB Radar and Transformer}, 
  year={2023},
  volume={11},
  number={},
  pages={15082-15095},
  doi={10.1109/ACCESS.2023.3244017}}

@inproceedings{rtpose,
  title={RT-Pose: A 4D Radar Tensor-Based 3D Human Pose Estimation and Localization Benchmark},
  author={Ho, Yuan-Hao and Cheng, Jen-Hao and Kuan, Sheng Yao and Jiang, Zhongyu and Chai, Wenhao and Huang, Hsiang-Wei and Lin, Chih-Lung and Hwang, Jenq-Neng},
  booktitle={European Conference on Computer Vision},
  pages={107--125},
  year={2024},
  organization={Springer}
}

@article{mmpose,
  title={mm-Pose: Real-time human skeletal posture estimation using mmWave radars and CNNs},
  author={Sengupta, Arindam and Jin, Feng and Zhang, Renyuan and Cao, Siyang},
  journal={IEEE Sensors Journal},
  volume={20},
  number={17},
  pages={10032--10044},
  year={2020},
  publisher={IEEE}
}

@inproceedings{hupr,
  title={Hupr: A benchmark for human pose estimation using millimeter wave radar},
  author={Lee, Shih-Po and Kini, Niraj Prakash and Peng, Wen-Hsiao and Ma, Ching-Wen and Hwang, Jenq-Neng},
  booktitle={Proceedings of the IEEE/CVF Winter Conference on Applications of Computer Vision},
  pages={5715--5724},
  year={2023}
}

@inproceedings{transhupr,
author    = {Niraj Prakash Kini and Ruey-Horng Shiue and ryan chandra and Wen-Hsiao Peng and Ching-Wen Ma and Jenq-Neng Hwang},
title     = {TransHuPR: Cross-View Fusion Transformer for Human Pose Estimation Using mmWave Radar},
booktitle = {35th British Machine Vision Conference 2024, {BMVC} 2024, Glasgow, UK, November 25-28, 2024},
publisher = {BMVA},
year      = {2024},
url       = {https://papers.bmvc2024.org/0686.pdf}
}

@inproceedings{ap_metric,
  title={Microsoft coco: Common objects in context},
  author={Lin, Tsung-Yi and Maire, Michael and Belongie, Serge and Hays, James and Perona, Pietro and Ramanan, Deva and Doll{\'a}r, Piotr and Zitnick, C Lawrence},
  booktitle={Computer vision--ECCV 2014: 13th European conference, zurich, Switzerland, September 6-12, 2014, proceedings, part v 13},
  pages={740--755},
  year={2014},
  organization={Springer}
}

@misc{radar_module,
  author       = {{Texas Instruments}},
  title        = {{IWR1843BOOST Evaluation Module}},
  howpublished = {\url{https://www.ti.com/tool/IWR1843BOOST}},
}

@article{transformer,
  title={Attention is all you need},
  author={Vaswani, Ashish and Shazeer, Noam and Parmar, Niki and Uszkoreit, Jakob and Jones, Llion and Gomez, Aidan N and Kaiser, {\L}ukasz and Polosukhin, Illia},
  journal={Advances in neural information processing systems},
  volume={30},
  year={2017}
}

@article{mamba,
  title={Mamba: Linear-time sequence modeling with selective state spaces},
  author={Gu, Albert and Dao, Tri},
  journal={arXiv preprint arXiv:2312.00752},
  year={2023}
}

@inproceedings{vim,
  title={Vision Mamba: Efficient Visual Representation Learning with Bidirectional State Space Model},
  author={Zhu, Lianghui and Liao, Bencheng and Zhang, Qian and Wang, Xinlong and Liu, Wenyu and Wang, Xinggang},
  booktitle={Forty-first International Conference on Machine Learning},
  year={2024}
}

@inproceedings{rfmamba,
  title={RFMamba: Frequency-Aware State Space Model for RF-Based Human-Centric Perception},
  author={Zhang, Rui and Geng, Ruixu and Li, Yadong and Song, Ruiyuan and Gong, Hanqin and Zhang, Dongheng and Hu, Yang and Chen, Yan},
  booktitle={The Thirteenth International Conference on Learning Representations},
  year={2025}
}

@inproceedings{DETR,
  title={End-to-end object detection with transformers},
  author={Carion, Nicolas and Massa, Francisco and Synnaeve, Gabriel and Usunier, Nicolas and Kirillov, Alexander and Zagoruyko, Sergey},
  booktitle={European conference on computer vision},
  pages={213--229},
  year={2020},
  organization={Springer}
}

@inproceedings{mnet,
  title={Rodnet: Radar object detection using cross-modal supervision},
  author={Wang, Yizhou and Jiang, Zhongyu and Gao, Xiangyu and Hwang, Jenq-Neng and Xing, Guanbin and Liu, Hui},
  booktitle={Proceedings of the IEEE/CVF Winter Conference on Applications of Computer Vision},
  pages={504--513},
  year={2021}
}

@inproceedings{smoothnet,
  title={Smoothnet: A plug-and-play network for refining human poses in videos},
  author={Zeng, Ailing and Yang, Lei and Ju, Xuan and Li, Jiefeng and Wang, Jianyi and Xu, Qiang},
  booktitle={European Conference on Computer Vision},
  pages={625--642},
  year={2022},
  organization={Springer}
}

@article{cpformer,
  title={CPFormer: End-to-End Multi-Person Human Pose Estimation From Raw Radar Cubes With Transformers},
  author={Chen, Lin and Wang, Guoli},
  journal={IEEE Sensors Journal},
  year={2025},
  publisher={IEEE}
}

@inproceedings{rfpose,
  title={Through-wall human pose estimation using radio signals},
  author={Zhao, Mingmin and Li, Tianhong and Abu Alsheikh, Mohammad and Tian, Yonglong and Zhao, Hang and Torralba, Antonio and Katabi, Dina},
  booktitle={Proceedings of the IEEE conference on computer vision and pattern recognition},
  pages={7356--7365},
  year={2018}
}

@inproceedings{videomamba,
  title={Videomamba: State space model for efficient video understanding},
  author={Li, Kunchang and Li, Xinhao and Wang, Yi and He, Yinan and Wang, Yali and Wang, Limin and Qiao, Yu},
  booktitle={European conference on computer vision},
  pages={237--255},
  year={2024},
  organization={Springer}
}

@inproceedings{tokenpose,
  title={Tokenpose: Learning keypoint tokens for human pose estimation},
  author={Li, Yanjie and Zhang, Shoukui and Wang, Zhicheng and Yang, Sen and Yang, Wankou and Xia, Shu-Tao and Zhou, Erjin},
  booktitle={Proceedings of the IEEE/CVF International conference on computer vision},
  pages={11313--11322},
  year={2021}
}

@inproceedings{mvdoppler,
  title={MVDoppler-Pose: Multi-Modal Multi-View mmWave Sensing for Long-Distance Self-Occluded Human Walking Pose Estimation},
  author={Choi, Jaeho and Hor, Soheil and Yang, Shubo and Arbabian, Amin},
  booktitle={Proceedings of the Computer Vision and Pattern Recognition Conference},
  pages={27750--27759},
  year={2025}
}

@article{probradarm3f,
  title={ProbRadarM3F: mmWave Radar-based Human Skeletal Pose Estimation with Probability Map Guided Multi-Format Feature Fusion},
  author={Zhu, Bing and He, Zixin and Xiong, Weiyi and Ding, Guanhua and Huang, Tao and Xiang, Wei},
  journal={IEEE Transactions on Aerospace and Electronic Systems},
  year={2025},
  publisher={IEEE}
}

@article{lv2025egohand,
  title={EgoHand: Ego-centric Hand Pose Estimation and Gesture Recognition with Head-mounted Millimeter-wave Radar and IMUs},
  author={Lv, Yizhe and Zhang, Tingting and Song, Yunpeng and Ding, Han and Han, Jinsong and Wang, Fei},
  journal={arXiv preprint arXiv:2501.13805},
  year={2025}
}

@inproceedings{wang2024gtpt,
  title={GTPT: Group-based token pruning transformer for efficient human pose estimation},
  author={Wang, Haonan and Liu, Jie and Tang, Jie and Wu, Gangshan and Xu, Bo and Chou, Yanbing and Wang, Yong},
  booktitle={European Conference on Computer Vision},
  pages={213--230},
  year={2024},
  organization={Springer}
}

@inproceedings{rahman2024mmvr,
  title={MMVR: Millimeter-wave multi-view radar dataset and benchmark for indoor perception},
  author={Rahman, M Mahbubur and Yataka, Ryoma and Kato, Sorachi and Wang, Pu and Li, Peizhao and Cardace, Adriano and Boufounos, Petros},
  booktitle={European Conference on Computer Vision},
  pages={306--322},
  year={2024},
  organization={Springer}
}

@article{yataka2024retr,
  title={RETR: Multi-view radar detection transformer for indoor perception},
  author={Yataka, Ryoma and Cardace, Adriano and Wang, Perry and Boufounos, Petros and Takahashi, Ryuhei},
  journal={Advances in Neural Information Processing Systems},
  volume={37},
  pages={19839--19869},
  year={2024}
}

@article{wu2024mmhpe,
  title={mmhpe: Robust multi-scale 3d human pose estimation using a single mmwave radar},
  author={Wu, Yingxiao and Jiang, Zhongmin and Ni, Haocheng and Mao, Changlin and Zhou, Zhiyuan and Wang, Wenxiang and Han, Jianping},
  journal={IEEE Internet of Things Journal},
  year={2024},
  publisher={IEEE}
}

@article{mueller2025radproposer,
  title={RadProPoser: A Framework for Human Pose Estimation with Uncertainty Quantification from Raw Radar Data},
  author={Mueller, Jonas Leo and Engel, Lukas and Dorschky, Eva and Krauss, Daniel and Ullmann, Ingrid and Vossiek, Martin and Eskofier, Bjoern M},
  journal={arXiv preprint arXiv:2508.03578},
  year={2025}
}

@inproceedings{huang2025one,
  title={One Snapshot is All You Need: A Generalized Method for mmWave Signal Generation},
  author={Huang, Teng and Ding, Han and Sun, Wenxin and Zhao, Cui and Wang, Ge and Wang, Fei and Zhao, Kun and Wang, Zhi and Xi, Wei},
  booktitle={IEEE INFOCOM 2025-IEEE Conference on Computer Communications},
  pages={1--10},
  year={2025},
  organization={IEEE}
}

@article{engel2025advanced,
  title={Advanced Millimeter Wave Radar-Based Human Pose Estimation Enabled by a Deep Learning Neural Network Trained With Optical Motion Capture Ground Truth Data},
  author={Engel, Lukas and Mueller, Jonas and Rendon, Eduardo Javier Feria and Dorschky, Eva and Krauss, Daniel and Ullmann, Ingrid and Eskofier, Bjoern M and Vossiek, Martin},
  journal={IEEE Journal of Microwaves},
  year={2025},
  publisher={IEEE}
}

@article{yan2024indoor,
  title={Indoor 3D Human Pose Estimation Using Single Millimeter-wave Radar And Conditional Generative Adversarial Network},
  author={Yan, Zhihang and Ni, Hongbo and Yu, Xiaoguang and Tan, Weiqi},
  journal={IEEE Sensors Journal},
  year={2024},
  publisher={IEEE}
}

@article{jin2024rodar,
  title={Rodar: Robust gesture recognition based on mmWave radar under human activity interference},
  author={Jin, Can and Meng, Xiangzhu and Li, Xuanheng and Wang, Jie and Pan, Miao and Fang, Yuguang},
  journal={IEEE Transactions on Mobile Computing},
  volume={23},
  number={12},
  pages={11735--11749},
  year={2024},
  publisher={IEEE}
}

@article{xie2023rpm,
  title={RPM 2.0: RF-based pose machines for multi-person 3D pose estimation},
  author={Xie, Chunyang and Zhang, Dongheng and Wu, Zhi and Yu, Cong and Hu, Yang and Chen, Yan},
  journal={IEEE Transactions on Circuits and Systems for Video Technology},
  volume={34},
  number={1},
  pages={490--503},
  year={2023},
  publisher={IEEE}
}

@inproceedings{jiao2025optimizing,
  title={Optimizing human pose estimation through focused human and joint regions},
  author={Jiao, Yingying and Wang, Zhigang and Liu, Zhenguang and Fan, Shaojing and Wu, Sifan and Wu, Zheqi and Xu, Zhuoyue},
  booktitle={Proceedings of the AAAI Conference on Artificial Intelligence},
  volume={39},
  number={4},
  pages={4102--4110},
  year={2025}
}

@inproceedings{liu2025tcpformer,
  title={Tcpformer: Learning temporal correlation with implicit pose proxy for 3d human pose estimation},
  author={Liu, Jiajie and Liu, Mengyuan and Liu, Hong and Li, Wenhao},
  booktitle={Proceedings of the AAAI Conference on Artificial Intelligence},
  volume={39},
  number={5},
  pages={5478--5486},
  year={2025}
}
}

\end{document}